\documentclass[aps,preprint,groupedaddress]{revtex4}
\usepackage{amsmath,amssymb,amsthm,mathrsfs,amsfonts,dsfont} 
\usepackage{graphicx}
\usepackage{times}
\usepackage{subfigure}
\usepackage{amsbsy}
\usepackage{bm}
\usepackage{algorithm}
\usepackage{algpseudocode}
\usepackage{algorithmicx}

\usepackage{xcolor}

\def\bx{\mathbf{x}}
\def\id{\mathbf{I}}
\def\bw{\mathbf{w}}
\def\by{\mathbf{y}}
\def\bs{\mathbf{s}}
\def\bb{\mathbf{b}}
\def\bz{\mathbf{z}}
\def\bbt{\tilde{\mathbf{b}}}
\def\bsgm{\boldsymbol{\sigma}}
\def\bmu{\boldsymbol{\mu}}

\begin{document}

\title{Solvable dynamical model of in-context learning}

\author{Ruoran Bai$^{1}$}
\author{Haiping Huang$^{1,2}$}
\email{huanghp7@mail.sysu.edu.cn}
\affiliation{$^{1}$PMI Lab, School of Physics,
Sun Yat-sen University, Guangzhou 510275, People's Republic of China}
\affiliation{$^{2}$Guangdong Provincial Key Laboratory of Magnetoelectric Physics and Devices,
Sun Yat-sen University, Guangzhou 510275, People's Republic of China}

\date{\today}

\begin{abstract}
Pretrained Transformer architectures display a surprising in-context learning ability without additional training. How the learning process guides the network to achieve this intriguing capability remains unsolved. In this study, we focus on a single-layer single-head linear attention model, where an empirical risk minimization is defined for learning. Given a finite number of training examples, we determine the exact solution of network-parameter dynamics, which is linked to the training error and generalization error, characterizing how memorization and generalization emerge from learning and how these transitions are related to curvatures of the empirical risk landscape. Furthermore, we calculate a data-dependent free energy, from which the training and generalization errors in the long time limit of learning dynamics can be well predicted. Our work adds insights toward understanding the mechanism of in-context learning especially from a learning dynamics perspective.

\end{abstract}

\maketitle

\section{Introduction}
Inspired by the study of attention mechanisms in the brain~\cite{Harris-2011,VA-2015,KV-2025}, an engineering generalization of the biological attention to the self-attention dominant in the current transformer computation has started a revolution in natural language processing first and recently the artificial general intelligence including superhuman capabilities in coding, mathematical reasoning, and document comprehension~\cite{Att-2017,AGI-2026}. The self-attention captures long-range dependencies between different parts of the input (e.g., image or text), trained with a simple cost of next-token prediction~\cite{Uni-2023}, establishing a computer science foundation of large language model (LLM)~\cite{Sparks-2023}. 
 
 As one of the astonishing abilities, the transformer is able to do in-context inference~\cite{LLM-2020}, i.e., the pre-trained transformer is able to accomplish previously-unseen complicated tasks by reasoning on a short prompt especially without a need for updating the model parameters. This remarkable property has recently captured intense research interest from theoretical machine learning and statistical physics communities~\cite{Xie-2021,Wu-2024,Ganguli-2024,Aky-2022,Von-2023,CP-2025,Li-2025,Zhang-2024,Yang-2025}. The theoretical efforts focused on a single-layer linear attention~\cite{Aky-2022,Von-2023,CP-2025, Li-2025, Zhang-2024}, which revealed that a sufficient pre-training task diversity triggers the emergence of in-context inference. This phase transition was first empirically confirmed in the in-context learning of a regression task~\cite{Ganguli-2024}, later analytically studied in the asymptotic high-dimensional limit~\cite{CP-2025}, and formulated recently in a disordered spin glass framework where the network parameters act as spins while training data act as quenched coupling between spins. Recent theoretical works start to focus on the learning dynamics of transformer in linear regression tasks, especially on the population loss (given infinite training prompts)~\cite{Zhang-2024,Zhang-2025} and on the neural scaling law in training time, depth and context length~\cite{Bord-2025}. A recent work derives non-asymptotic closed-form characterization of the learning dynamics for vector-valued regression~\cite{Exact-2025}.
 
Based on our previous work of statistical physics analysis of in-context learning (ICL), we continue to explore the learning dynamics of an empirical risk minimization in this work. We first describe the model setting which allows a closed-form solution of the network parameter dynamics, from which we can predict the behavior of individual concrete learning trajectory and related test performance on unseen linear regression examples. We also derive a data-dependent statistical mechanics formalism to explain the stationary limit of the learning dynamics, supported by numerical experiments of single instance learning. Furthermore, we simplify our analytic results in a specific setting, e.g., setting a particular set of network parameters to zero, thereby isolating the key factor in the form of a data-dependent eigenvalue spectrum.

\section{Linear attention model}
\label{LATT}
Given an input matrix $X\in\mathbb{R}^{D\times N}$, a single-head single-layer softmax-based attention model is formulated as follows,
\begin{align}
 & \text{Atten}(K,Q,V) =  V\text{softmax}\left( \frac{Q^\top K}{\sqrt{ D }} \right)\\
 &K = W_{K}X;\;Q=W_{Q}X;\;V=W_{V}X,
\end{align}
where $W_{\{Q,K,V\}}$ indicates the query, key or value matrix of dimension $D\times D$, respectively. $D$ is also called the embedding dimension of token $\mathbf{x}_i\{i=1,\ldots,N\}$.  All vectors in this paper are defined in a column form, i.e., $\bx\in\mathbb{R}^{D\times 1}$. The above transformer block outputs an equal-size matrix 
$\mathbf{Y}=\text{Atten}(K,Q,V)$. The softmax operation is normalized for each column, representing a weight score for different key vectors (transformed tokens). 
To facilitate the following theoretical analysis, we replace the softmax operation by a linear attention, and merge the key and query projection matrices into a single parameter matrix $W_Q^\top W_K=W$, and set the value projection matrix $W_V=\mathbf{I}$. This results in a simplified single-layer linear attention model:
\begin{equation}
\mathrm{Atten}(X,W)=\frac{1}{N}\,X X^\top W X,
\end{equation}
where $W\in\mathbb{R}^{D\times D}$ is trainable. 

Next, we design the learning task. We consider the following input structure:
\begin{equation}
X = \begin{bmatrix}
    \bx_1 & \bx_2 & \dots & \bx_N & \bx_q \\
    y_1 & y_2 & \dots & y_N & 0
\end{bmatrix},
\end{equation}
where in each column, the last element serves as the label $y_\ell=(\mathbf{w}^\mu)^\top\mathbf{x}_\ell$. In particular, for the last column, only the query vector $\mathbf{x}_q$ is shown, and the answer label is masked.
Each vector $\mathbf{x}\in\mathbb{R}^{D}$, are sampled from an i.i.d. distribution $\mathcal{N}(0,\id/D)$, and for each task $\mu$, the input $X$ is generated using the same task vector $\bw$, with each component generated independently from standard
Gaussian distribution. Thanks to the structure of $X$, we call the task as a linear regression in-context learning, and thus the token embedding dimension is now $D+1$, with the extra dimension for the label. 

The pretraining requires that the model predicts the label for the last question $\bx_q$. The model's scalar prediction is the $(D{+}1,N{+}1)$ entry of the attention output,
\begin{equation}
\hat{y} = \bigl[\tfrac{1}{N} X X^\top W X\bigr]_{D+1,\,N+1}.
\end{equation}
The pretraining cost function is given below:
\begin{equation}\label{error}
\mathcal{L}(W)=\frac{1}{2P}\sum_{\mu=1}^P\bigl(y^\mu-\hat{y}^\mu\bigr)^{2}+\frac{\lambda}{2}\lVert W \rVert^2,
\end{equation}
where $\lambda\ge 0$ controls $\ell_2$ regularization (weight decay), $P$ is the number of training examples with different task vectors but all belonging to the same linear regression problem. To sum up, the key model parameters for this empirical risk minimization are set by the task diversity $P$, context length $N$ and input dimension $D$. The size of weight matrix $W$ is determined by $(D+1)^2$, whose actual size for learning or inference is smaller (to be explained later). To keep both terms in Eq.~\eqref{error} order of $\mathcal{O}(1)$, $\lambda$ may be scaled by $1/D^3$ if $D$ is a large value.

To proceed, we first compute $\frac{1}{N}X^\mu (X^\mu)^\top$, where $X^\mu$ is organized into a block matrix:
\begin{equation}
X = \begin{bmatrix}
X_{0} & \bx_{q} \\
\by_{0}^\top & 0
\end{bmatrix}
\end{equation}
where $\mathbf{y}_0=X_0^\top\bw^\mu\in\mathbb{R}^N$. We then obtain:
\begin{equation} \label{eq:XXT}
    \mathbf{XX^\top} = 
    \left[\begin{array}{cc}
        \mathbf{X}_0 & \bx_q \\
        \by_0^\top  & 0 
    \end{array}\right]
    \left[\begin{array}{cc}
        \mathbf{X}_0^\top & \by_0 \\
        \bx_q^\top  & 0 
    \end{array}\right]
    = 
    \left[\begin{array}{cc}
        \mathbf{X}_0 \mathbf{X}_0^\top + \bx_q \bx_q^\top & \mathbf{X}_0 \by_0  \\
        \by_0^\top \mathbf{X}_0^\top  & \by_0^\top \by_0
    \end{array}\right] .
\end{equation}
Therefore, we define $\Gamma^\mu=\frac{1}{N}X^\mu (X^\mu)^\top$, and obtain a compact form:
\begin{equation}
\Gamma^\mu = \begin{bmatrix}
\Sigma_x+\frac{1}{N}\bx_q\bx_q^\top & \Sigma_x\bw^\mu \\
(\bw^\mu)^\top\Sigma_x & (\bw^\mu)^\top\Sigma_x\bw^\mu
\end{bmatrix},
\end{equation}
where $\Sigma_x=\frac{1}{N}X_0X_0^\top$, and $X_0$ depends on the example index $\mu$ (in the following this index is implicitly included).

Next, we write the weight matrix into a block form, 
\begin{equation}
W = \begin{bmatrix}
W_{11} & W_{12} \\
W_{21} & W_{22}
\end{bmatrix}.
\end{equation}
 Thus, we get the product $WX$:
\begin{equation}
    \mathbf{WX} = 
    \left[\begin{array}{cc}
        \mathbf{W}_{11} & \mathbf{W}_{12} \\
        \mathbf{W}_{21} & \mathbf{W}_{22} 
    \end{array}\right] 
    \left[\begin{array}{cc}
        \mathbf{X}_0 & \bx_q\\
        \by_0^\top  & 0
    \end{array}\right]
    = 
    \left[\begin{array}{cc}
        \mathbf{W}_{11} \mathbf{X}_0 + \mathbf{W}_{12}\by_0^\top & \mathbf{W}_{11} \bx_q \\
        \mathbf{W}_{21} \mathbf{X}_0 + \mathbf{W}_{22} \by_0^\top & \mathbf{W}_{21} \bx_q
    \end{array}\right] .
\end{equation}
We finally compute the actual prediction of the linear attention model:
\begin{align}\label{predout}
\hat{y}^\mu&=\left[\Gamma^\mu WX\right]_{D+1,N+1}\\
&=(\bw^\mu)^\top\Sigma_x[W_{11}+\bw^\mu W_{21}]\bx_q\\
&=\mathrm{vec}\left(
\begin{bmatrix}
\Sigma_x\bw^\mu\bx_q^\top \\
(\bw^\mu)^\top\Sigma_x\bw^\mu \bx_q^\top
\end{bmatrix}\right)^{\top}
\mathrm{vec}\left(
\begin{bmatrix}
W_{11}\\
W_{21}
\end{bmatrix}\right) = \bs_\mu^\top \bsgm,
\end{align}
where $\bs_\mu$ and $\bsgm$ are the flattened forms of the data matrix and weight matrix, shown respectively below.
\begin{equation}
\bs_\mu=\mathrm{vec}\left(
\begin{bmatrix}
\Sigma_x\bw^\mu\bx_q^\top \\
(\bw^\mu)^\top\Sigma_x\bw^\mu \bx_q^\top
\end{bmatrix}\right)\in\mathbb{R}^{D(D+1)},
\qquad
\bsgm=\mathrm{vec}\left(
\begin{bmatrix}
W_{11}\\
W_{21}
\end{bmatrix}\right)\in\mathbb{R}^{D(D+1)}.
\end{equation}
Given this mapping, we can interpret $\bsgm$ as spin vector, while the data-dependent elements can be organized into quenched disorder by inserting the above reorganized prediction into the mean-squared error in Eq.~\eqref{error}. Then we can get the spherical spin model mapping of the ICL in transformer, whose statistical properties have been studied in our recent work~\cite{Li-2025}.
We finally remark that if $D$ is fixed but $N$ is large, then we have $\Sigma_x\to\id_D/D$. In this setting, a perfect ICL solution is given by $W_{11}=D\id_D$ and $W_{21}=0_D$, while $W_{12}$ and $W_{22}$ are irrelevant for the task.
 In this work, we focus on the learning dynamics and the cost function becomes now a function of $\bsgm$, i.e., $\mathcal{L}(\bsgm)$,  which shall be detailed in the following sections.

\section{Solvable dynamics}
\label{sDyn}
In this section, we study the gradient flow of the pretraining, which is shown as follows.
\begin{align}\label{gflow}
 \dot{\bsgm}& = - \frac{\partial \mathcal{L}(\bsgm)}{\partial \bsgm} \\
 &= \frac{1}{P}\bigl(S\by-SS^\top\bsgm\bigr)-\lambda \bsgm\\
  & = \frac{1}{P}S\by-(\frac{1}{P}SS^\top+\lambda\id)\bsgm, 
\end{align}
where $\by=[y^1,\ldots,y^P]^\top$, $S=[\bs_{1},\ldots,\bs_{P}]\in\mathbb{R}^{M\times P}$, $M=D(D+1)$ is the total number of weight parameters to be learned.

To proceed, we write $A=\frac{1}{P}SS^\top+\lambda\id$ and $\bb=\frac{1}{P}S\by$. Then we recast the learning dynamics into the following linear ordinary differential equation:
\begin{equation}\label{leq}
\frac{d\bsgm}{dt}\;=\;-A\bsgm + \bb
\quad\Longrightarrow\quad
\bsgm(t)=e^{-At}\bsgm(0)- A^{-1}\bigl(e^{-At}-\id\bigr)\bb.
\end{equation}
Clearly, when $t\to\infty$, a stationary solution (gradient-flow solution) is reached as $\bsgm^*=A^{-1}\bb$. Because $A$ is a symmetric matrix, we can do eigenvalue decomposition as $A=V\Lambda V^\top$, where $V$ is an orthogonal matrix satisfying $V^{-1}=V^\top$. 
This requires us to project $\bsgm$ and $\bb$ onto the eigenbasis $V$. Therefore, we introduce the projections as $\bz=V^\top\bsgm$ and $\bbt=V^\top\bb$. We finally arrive at the following solution of the learning dynamics:
 \begin{equation}\label{zeq}
\bz(t)=e^{-\Lambda t}\bz(0)-\Lambda^{-1}\bigl(e^{-\Lambda t}-\id\bigr)\bbt.
\end{equation}

Next, we calculate the time-dependent training error as follows.
\begin{align}
E_t
&=\frac{1}{2P}\bigl\lVert \by-S^\top\bsgm\bigr\rVert^2+\frac{\lambda}{2}\lVert \bsgm\rVert^2 \\
&=\frac{1}{2P}\left(\by^\top\by-2\by^\top S^\top\bsgm+\bsgm^\top SS^\top\bsgm\right)+\frac{\lambda}{2}\bsgm^\top\bsgm\\
&=\frac{1}{2P}\left(\by^\top\by-2P\bbt^\top\bz+\bz^\top\tilde{\Lambda}\bz\right)+\frac{\lambda}{2}\bz^\top\bz,
\end{align}
where $SS^\top=V\tilde{\Lambda} V^\top$, $\{\tilde{\Lambda}_i\}$ are the eigenvalues of $SS^\top$, while the eigenvalues in Eq.~\eqref{zeq} are related to the eigenvalues of $SS^\top$ by $\Lambda_i=\frac{1}{P}\tilde{\Lambda}_i+\lambda$. $SS^\top$ shares the same eigenvectors with $A$ in Eq.~\eqref{leq}.

Finally, we can calculate the test error in a similar way. We write the test data matrix $S_{\rm test}\in\mathbb{R}^{M\times P_{\rm test}}$, where $M$ is the size of model (parameters), and $\by_{\rm test}\in\mathbb{R}^{P_{\rm test}}$. The empirical test performance is characterized by the following formula:
\begin{equation}
E_g=\frac{1}{2}\left(a_{\rm te}-2(V^\top g_{\rm te})^\top\bz+\bz^\top V^\top Q_{\rm te}V\bz\right),
\end{equation}
where $a_{\rm te}\equiv\by_{\rm test}^\top\by_{\rm test}/P_{\rm test}$, $Q_{\rm te}\equiv S_{\rm test}S^\top_{\rm test}/P_{\rm test}$,
$g_{\rm te}\equiv S_{\rm test}\by_{\rm test}/P_{\rm test}$.

\section{Data-dependent free energy}
\label{dFree}
For pretraining the transformer, we consider the following full-batch gradient descent:
\begin{equation}\label{LEq}
 \dot{\bsgm} = -\gamma \frac{\partial \mathcal{L}(\bsgm)}{\partial \bsgm}+\eta(t),
\end{equation}
where $\gamma$ denotes the learning rate, and $\boldsymbol{\eta}(t)$ is a zero mean white noise vector, whose correlation is determined by $\langle\eta_i(t)\eta_j(t')\rangle=2T\gamma\delta_{ij}\delta(t-t')$. The above learning dynamics are also called Langevin dynamics and in the long time limit the statistics of the network parameters are subject to the following Bayesian posterior of the model~\cite{Risken-1996}:
\begin{align}
 P(\bsgm |\mathcal{D})=\frac{1}{Z}\exp\left[-\beta\cal{L}(\bsgm)\right],
\end{align}
where $\beta=\frac{1}{T}$, $\cal{D}$ indicates the training dataset $\{X^\mu,y^\mu\}_{\mu=1}^{P}$, and $Z$ is the data-dependent partition function. The partition function can be computed as follows.
\begin{subequations}\label{pfun}
\begin{align}
Z&=\int d\bsgm\,\exp\Bigl(-\frac{\beta}{2P}\bigl\lVert \by-S^\top\bsgm\bigr\rVert^2-\frac{\lambda\beta}{2}\lVert\bsgm\rVert^2\Bigr) \\
& = \int d\bsgm \exp(-\frac{\beta}{2P} \left( \by-S^\top\bsgm \right)^\top\left(\by-S^\top\bsgm \right)-\frac{\lambda\beta}{2}\bsgm^\top\bsgm)\\
& = \exp(-\frac{\beta}{2P} \by^\top\by)\int d\bsgm \exp\left(\frac{\beta}{P} \by^\top S^\top\bsgm -\frac{1}{2} \bsgm ^\top ( \frac{\beta}{P}S S^\top+\lambda\beta I )\bsgm\right) \\
&=\sqrt{\frac{(2\pi)^{M}}{\det\bigl(\tfrac{\beta}{P}S S^\top+\lambda\beta I\bigr)}}
\exp\Bigl(-\frac{\beta}{2P}\by^\top y+\frac{\beta}{2P}\,\by^\top S^\top(S S^\top+P\lambda I)^{-1}S\by\Bigr)\\
&=\sqrt{\frac{(2\pi)^{M}}{\det\bigl(\tfrac{\beta}{P}S S^\top+\lambda\beta I\bigr)}}
\exp\Bigl(-\frac{\beta}{2P}\by^\top \by+\frac{\beta}{2P}\,\by^\top (\id+P\lambda (S^{\top}S)^{-1})^{-1}\by\Bigr),
\end{align}
\end{subequations}
where the last equality employs $S^\top \left(S S^\top+a \id_M \right)^{-1}S=K(K+a\id_P)^{-1}=\id_P-a(K+a\id_P)^{-1}$, $a=P\lambda$, and $K=S^\top S$. Hence, the partition function can thus be simplified to the following result:
\begin{align}
Z =  \sqrt{ \frac{(2\pi)^{M}}{\det\left( \frac{\beta}{P}S S^\top+\lambda\beta \id \right)} } \exp(-\frac{1}{2} \by^\top C^{-1} \by),
\end{align}
where $C=\frac{1}{\lambda\beta}S^\top S+\frac{P}{\beta}\id$. Note that here we assume $\lambda>0$, but the above $Z$ is not applicable to the case of $\lambda=0$ and $P<M$. We do not focus on this regime with the data-dependent prediction in this section.

We next derive the equilibrium training error from the log-partition-function.
\begin{subequations}
\begin{align}
 \langle E_t\rangle_{\sigma} &=-\frac{\partial \ln Z}{\partial \beta}\\
 & = -\frac{\partial\left( -1/2 \log \det\left( \frac{\beta}{P}S S^\top+\lambda\beta I \right) - 1/2 \by^\top C^{-1}\by \right)}{\partial \beta}\\
 & = \frac{1}{2} \frac{\partial \log \det \left( \frac{\beta}{P}S S^\top+\lambda\beta I \right) }{\partial \beta}+\frac{1}{2} \by^\top \frac{\partial C^{-1}}{\partial \beta}\by\\
 &=\frac{M+\by^\top C^{-1}\by}{2\beta},
\end{align}
\end{subequations}
where to arrive at the last equality, we have used $\frac{1}{\det A}\frac{\partial \det(A)}{\partial x}= \operatorname{Tr}(A^{-1}\frac{\partial A}{\partial x})$ and 
$\frac{\partial C^{-1}}{\partial \beta}  = -C^{-1} \frac{\partial C}{\partial \beta}\cdot C^{-1}$.

Finally, we derive the expected test error over the Bayesian posterior $P(\bsgm)$. First, we notice that the posterior can be written into the multivariate Gaussian distribution:
\begin{subequations}
\begin{align}
P(\bsgm)&\propto \exp(-\beta \mathcal{L}(\sigma)) =  \exp(-\frac{\beta}{2P} \left( \by-S^\top\bsgm \right)^\top\left( \by-S^\top\bsgm \right)-\frac{\lambda\beta}{2}\bsgm^\top\bsgm)\\
& = \exp(-\frac{1}{2} \bsgm ^\top ( \frac{\beta}{P}S S^\top+\lambda\beta \id )\bsgm+\frac{\beta}{P} \by^\top S^\top\bsgm )\\
& = \exp( -\frac{1}{2}(\bsgm-\bmu)^\top \Sigma^{-1}(\bsgm-\bmu) ),
\end{align}
\end{subequations}
where the mean and covariance are $\bmu=\frac{\beta}{P}  \left( \frac{\beta}{P}S S^\top+\lambda\beta \id \right)^{-1}S \by$ and
  $\Sigma=\left( \frac{\beta}{P}S S^\top+\lambda\beta \id \right)^{-1}$, respectively.
  
  For a new test dataset $(S^{\mathrm{test}},\by^{\mathrm{test}})$, the test error obtained after training is given by:
\begin{subequations}
\begin{align}
\langle E_g\rangle&\equiv\mathbb{E}_{\bsgm\sim \mathcal{N}(\bmu,\Sigma)}\bigl[E_g(\bsgm)\bigr]
=\frac{1}{2P_{\mathrm{test}}}\,\mathbb{E}_{\bsgm\sim \mathcal{N}(\bmu,\Sigma)}\bigl[\lVert \by_{\mathrm{test}}-S_{\mathrm{test}}^\top \bsgm\rVert^2\bigr] \\
&=\frac{1}{2P_{\rm test}}\mathbb{E}_{\bsgm\sim \mathcal{N}(\bmu,\Sigma)}[\by_{\rm test}^\top \by^{\rm test}-2\by_{\rm test}^\top S_{\rm test}^\top\bsgm+\bsgm^\top S_{\rm test}S_{\rm test}^\top \bsgm] \\
& = \frac{1}{2P_{\rm test}}\left(\by_{\rm test}^\top \by^{\rm test}-2\by_{\rm test}^\top S_{\rm test}^\top\bmu+\operatorname{Tr}(S_{\rm test}S_{\rm test}^\top\Sigma)+\bmu^\top S_{\rm test}S_{\rm test}^\top\bmu\right),
\end{align}
\end{subequations}
where the final two terms in the last line are obtained by using $\mathbb{E}[x^\top Ax]=\mathbb{E}[\operatorname{Tr}(x^\top A x)]=\mathbb{E}[\operatorname{Tr}(Axx^\top)]=\operatorname{Tr}[A(\Sigma+\bmu \bmu^\top)]=\operatorname{Tr}(A\Sigma)+\bmu^\top A\bmu$. Finally, the entire expression can be reorganized into the following compact form:
\begin{align}
\langle E_g\rangle\equiv\mathbb{E}_{\bsgm\sim \mathcal{N}(\bmu,\Sigma)}\bigl[E_g(\bsgm)\bigr]
&=\frac{1}{2P_{\mathrm{test}}}\,\Bigl(\bigl\lVert \by_{\mathrm{test}}-S_{\mathrm{test}}^\top \bmu\bigr\rVert^2+\mathrm{Tr}\bigl(S_{\mathrm{test}} S_{\mathrm{test}}^\top \Sigma\bigr)\Bigr).
\end{align}
In the zero temperature limit, the second term in the bracket vanishes.

\section{Eigenspectra of the Hessian matrix}
\label{spe}
 Note that a key quantity determining the learning dynamics is the matrix $H=\frac{1}{P}SS^\top$, whose size is $M\times M$, $M=D(D+1)$ in our setting and elements $H_{ij}=\frac{1}{P}\sum_\mu S_{i}^\mu S_{j}^\mu$. $H$ is the very Hessian matrix of the equivalent spherical spin model (without regularization) of the in-context learning in our previous work~\cite{Li-2025}. The fraction of zero eigenvalues determines the landscape shape of our optimization, which is expected to affect the generalization performance. 
 
 According to the arguments made in previous works~\cite{Zhang-2024,Zhang-2025}, we set $W_{21}=\mathbf{0}$ for an additional analysis, as this zero solution is also observed in
our previous work~\cite{Li-2025}. The rationale can also be found in Eq.~\eqref{predout} when the linear regression can be perfectly inferred. Given a specified $\mu$, $S_{i}^\mu$ is a vectorization of
the matrix $\Sigma_x\bw^\mu\bx_q^\top$. The mean vanishes due to the i.i.d. setting of regression vector $\bx$ and
 task vector $\bw$. The covariance can also be determined as follows:
 \begin{align}\label{hess}
 \langle S_\ell^\mu S_{\ell'}^\nu\rangle&=\langle [\Sigma_x\bw^\mu]_i\bx_{q,j}\times[\Sigma_x\bw^\nu]_{i'}\bx_{q,j'}\rangle\\
 &=\frac{1}{D}\delta_{j,j'}\delta_{\mu,\nu}\langle(\Sigma_x)^2_{ii'}\rangle\\
 &=\frac{\kappa_D}{D^3}\delta_{i,i'}\delta_{j,j'}\delta_{\mu,\nu},
 \end{align}
where $\kappa_D=1+\frac{D+1}{N}$, $\ell=(i,j)$ and $\ell'=(i',j')$ due to the vectorization of a matrix, and to derive the last equality, the Wick's theorem is used.  We conclude that in this simplified case, $\Gamma=\frac{D^3}{\kappa_D}H$ is a random matrix whose eigenvalue spectrum can be determined by the Marchenko-Pastur law, according to the previous work~\cite{Zhou-2021}, which we shall verify by comparing the prediction with numerical simulations.

 \begin{figure}
    \centering
    \includegraphics[width=0.9\textwidth]{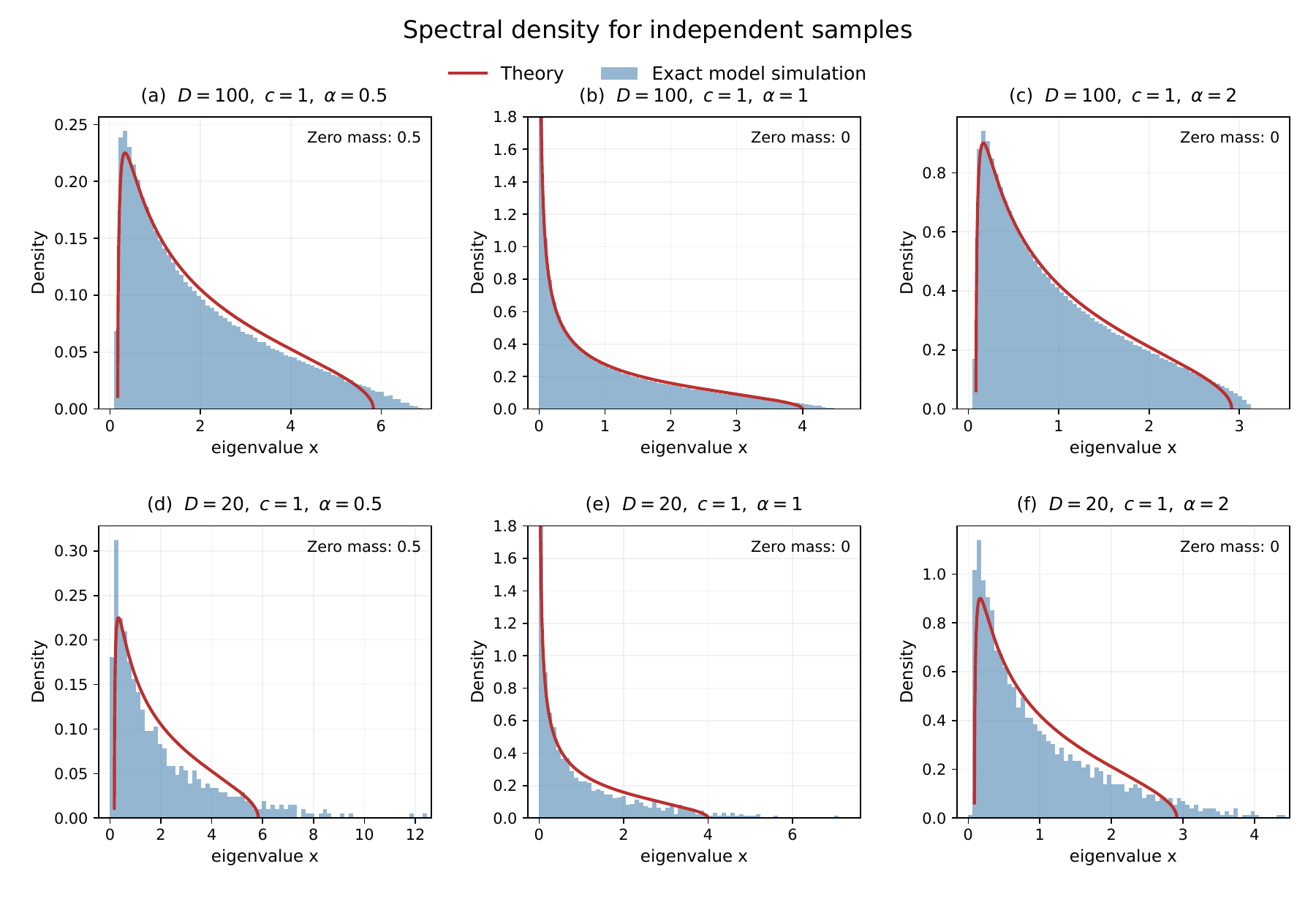}
    \caption{Eigenspectra of the rescaled Hessian matrix. Blue histograms: three independent datasets for each $\alpha$, with $D = N = 100$ (a-c) and $D=N=20$ (d-f). Red curves: Eq.~\eqref{spect}. The $\alpha= 0.5$ histogram excludes the zero eigenvalues but is normalized by all eigenvalues, so its area is $0.5$. The zero atom is stated separately. The $\alpha = 1$ density diverges at the origin; its vertical axis is
truncated to show the rest of the spectrum.}
        \label{fig0}
\end{figure}

To get the asymptotic spectral density of the matrix $\Gamma$, we assume $P=\alpha D^2$, $N=cD$. Therefore, the eigenvalue distribution is given by the following formula~\cite{Zhou-2021}:
\begin{equation}\label{spect}
\rho(x)=(1-\alpha)_{+}\delta(x)+\frac{\sqrt{(\ell^2x_{+}-x)(x-\ell^2x_{-})}}{2\pi\ell^2x}\id_{[x_{-},x_{+}]}(x),
\end{equation}
where $(y)_{+}=\operatorname{max}(0,y)$, $\id_{[x_{-},x_{+}]}(x)$ returns one if $x$ falls within the range $[\ell^2x_{-},\ell^2x_{+}]$ and zero otherwise, $x_{\pm}=(\sqrt{\alpha}\pm1)^2$, and $\ell^2=\frac{1}{\alpha}$. The results are shown in Fig.~\ref{fig0}, which demonstrates that $\alpha$ controls the fraction of zero eigenvalues, while with increasing the value of $D$ the agreement with the Marchenko-Pastur law is expected. We study the learning performance under this setting in the end of the next section.

\section{Theory corroborated by experiments}
\begin{figure}
    \centering
    \includegraphics[width=0.9\textwidth]{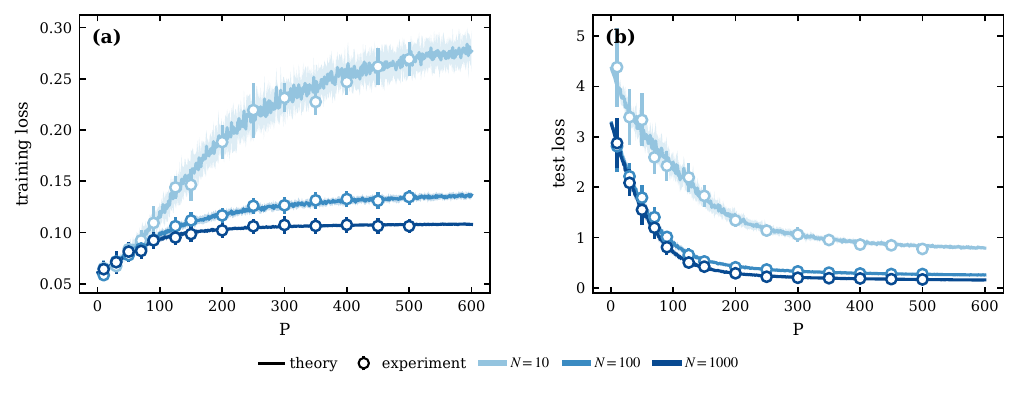}
    \caption{Stationary training and test losses as a function of the number of training prompts $P$ for different context lengths $N$, with fixed input dimension $D= 10$. Solid curves denote our theoretical predictions from the free-energy calculation in Sec.~\ref{dFree}, while markers represent numerical estimates from Langevin dynamics simulations [see Eq.~\eqref{LEq}]. In the simulations, we set $\beta=10^3, \lambda=0.01$, and $\gamma=1$. After an equilibration period of $10^5$ steps, we sample $20$ points separated by $5000$ steps and average them to estimate the final loss. (a) As the task diversity increases, the stationary training loss increases for all values of $N$. The smaller the value of $N$, the faster the training loss increases.
    (b) As the task diversity increases, the stationary test loss decreases for all values of $N$. However, for a given task diversity, smaller values of $N$ consistently yield a higher test loss. }
        \label{fig1}
\end{figure}

\begin{figure}
    \centering
    \includegraphics[width=0.9\textwidth]{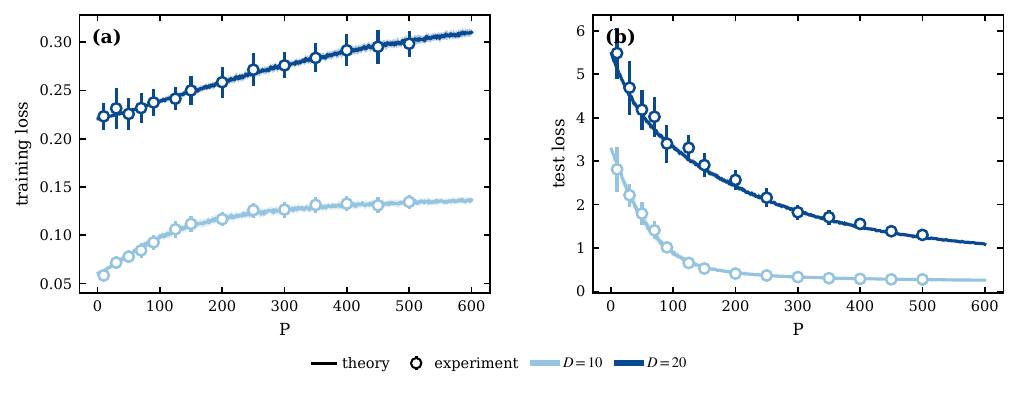}
    \caption{Stationary training and test losses as a function of the number of training prompts $P$ for different input dimensions $D$, with fixed context length $N = 100$. Solid curves denote the theoretical predictions from the free-energy calculation in Sec.~\ref{dFree}, while markers represent numerical estimates from Langevin dynamics simulations [Eq.~\eqref{LEq}]. In simulations, we set $\beta=10^3$ and $\lambda=0.01$, and $\gamma=1$. (a) The training loss increases with $P$ and is larger for a higher-dimensional input. (b) The test loss decreases as $P$ grows, but a larger $D$ requires more task diversity to reach a comparable generalization performance.}
        \label{fig2}
\end{figure}

\begin{figure}
    \centering
    \includegraphics[width=0.9\textwidth]{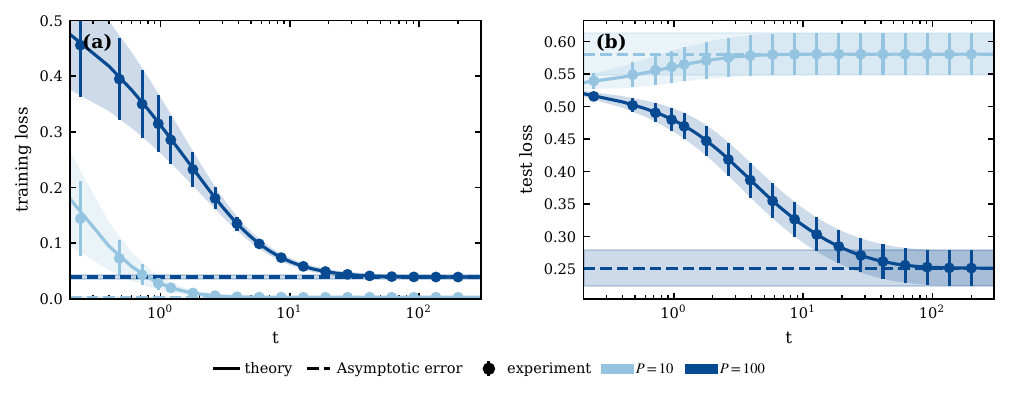}
    \caption{Gradient-flow dynamics of the training and test losses for task diversity $P=10$ and $100$, with $D=10$ and $N=100$. The  regularization parameter $\lambda=0.01$. Solid curves represent the analytical gradient-flow solution with the shaded regions indicating the theoretical uncertainty (see Eq.~\eqref{gflow} in Sec.~\ref{sDyn} in the main text), while markers represent gradient descent simulations. Asymptotic error means the stationary limit of the error predicted by the zero-temperature free energy analysis in Sec.~\ref{dFree}. Note that the training error includes the weight-decay term, while the test error does not include that term. (a) Training losses rapidly converge to their stationary value for both small and large $P$. (b) Test losses exhibit qualitatively different behavior depending on the task diversity: for the small $P$, the model exhibits memorization-like behavior and the test loss increases, whereas for the larger $P$, the test loss decreases monotonically, indicating the emergence of in-context generalization.}
        \label{fig3}
\end{figure}

First, we compare the data-dependent theory and the numerical simulations (in all simulations, we used zero-weight initialization and $\delta t=0.002$ unless otherwise stated), as shown in Fig.~\ref{fig1}. As the diversity increases, the training error increases, more rapidly for short context length. The number of training examples can be thought of as a learning complexity, e.g., fitting many more data points becomes hard. This phenomenon was also observed in training a single-layer neural network~\cite{Haim-2020}.  However, with increasing the task diversity, the generalization performance gets better, especially for a long context length, which coincides with our previous work~\cite{Li-2025}. Here, our theoretical results agree very well with the numerical Langevin learning dynamics. Because our theory depends on specified training dataset, there exists fluctuation related to the data choice. We also consider different levels of model complexity, characterized by the input dimension $D$ such that the the number of model parameters equals to $M=D(D+1)$. As revealed in Fig.~\ref{fig2}, more complex model requires more data examples to reach a comparable test loss to less complex model. The theory agrees with numerical simulations as well. 

Next, we study how the training and test losses evolve with training steps (see Fig.~\ref{fig3}). For the training loss, we observe a monotonically decreasing behavior, regardless of the task diversity. However, regarding the test loss, the behavior can be completely different depending on the task diversity. For a small value of $P$, the training can not improve the generalization performance but make the generalization even worse. Nevertheless, the training error is close to zero, demonstrating a perfect memorization of the training prompts. The experimental results are corroborated by theory, especially the asymptotic limit of $t\to\infty$ is captured by the data-dependent statistical mechanics theory in Sec.~\ref{dFree}.

\begin{figure}
    \centering
    \includegraphics[width=0.9\textwidth]{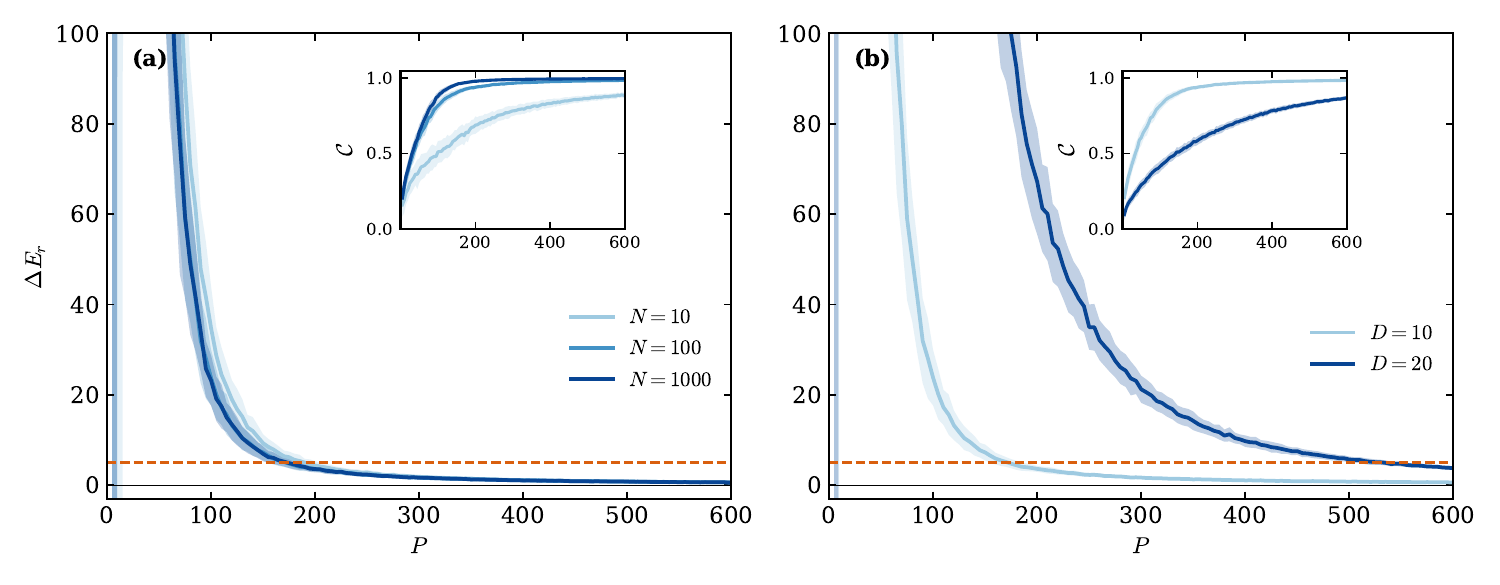}
    \caption{Dependence of the relative generalization gap $\Delta E_{r}$ and the contrast ratio $\mathcal{C}$ on task diversity $P$. The results are obtained by taking the long-time limit of the analytical gradient-flow solution derived in Sec.~\ref{sDyn} with the regularization strength $\lambda = 0.01$. The main panels show the generalization gap, while the insets show the contrast ratio. Shaded bands are $\pm1$ s.d. over $100$ training sets per $P$. Different color intensities correspond to different values of $N$ (or $D$). The relative gap can be slightly negative at large $P$ because $E_t$ includes the weight-decay term. (a) Results for different context lengths $N$ at fixed $D=10$. (b) Results for different input dimensions $D$ at fixed $N=100$. The dashed line of $\Delta E_r=5$ identifies a rough estimate of the value of $P$ separating memorization from generalization. }
        \label{fig4}
\end{figure}

The memorization effects motivate us to propose an indicator, namely the relative generalization gap $\Delta E_{r}=\frac{\langle E_g\rangle-\langle E_t\rangle}{\langle E_t\rangle}$. Note that here we remove the weight-decay term in computing $\langle E_t\rangle$ in order to be parallel to $\langle E_g\rangle$. In the case of a smaller $E_t$ and the corresponding larger $E_g$, the relative generalization gap will become very large; when the model generalizes, this gap will come to a relatively small level, at least of the order $\mathcal{O}(1)$. In addition, we also follow our previous work~\cite{Li-2025} to define the contrast ratio between the learned matrix $W_{11}$, and the identity matrix, which is a solution of perfect generalization except for a prefactor $D$, see  Eq.~\eqref{predout} when $N$ is independently large~\cite{Li-2025}. More precisely, $\mathcal{C}=\langle U,I_D\rangle/(\lVert U\rVert_2\lVert I_D\rVert_2)$ where $U$ indicates the non-zero transformer matrix $W_{11}/D$, $\langle A,B\rangle\equiv\sum_{ij}A_{ij}B_{ij}$, $\lVert A\rVert_2$ denotes a matrix Frobenius norm, and $I_D$ denotes a $D\times D$ identity matrix. As illustrated in Fig.~\ref{fig4}, a memorization phase is clearly separated from a generalization phase.
 Note that this transition is not in the thermodynamic sense, as we do not focus on the asymptotic limit of network sizes and data sizes.

\begin{figure}
    \centering
    \includegraphics[width=0.9\textwidth]{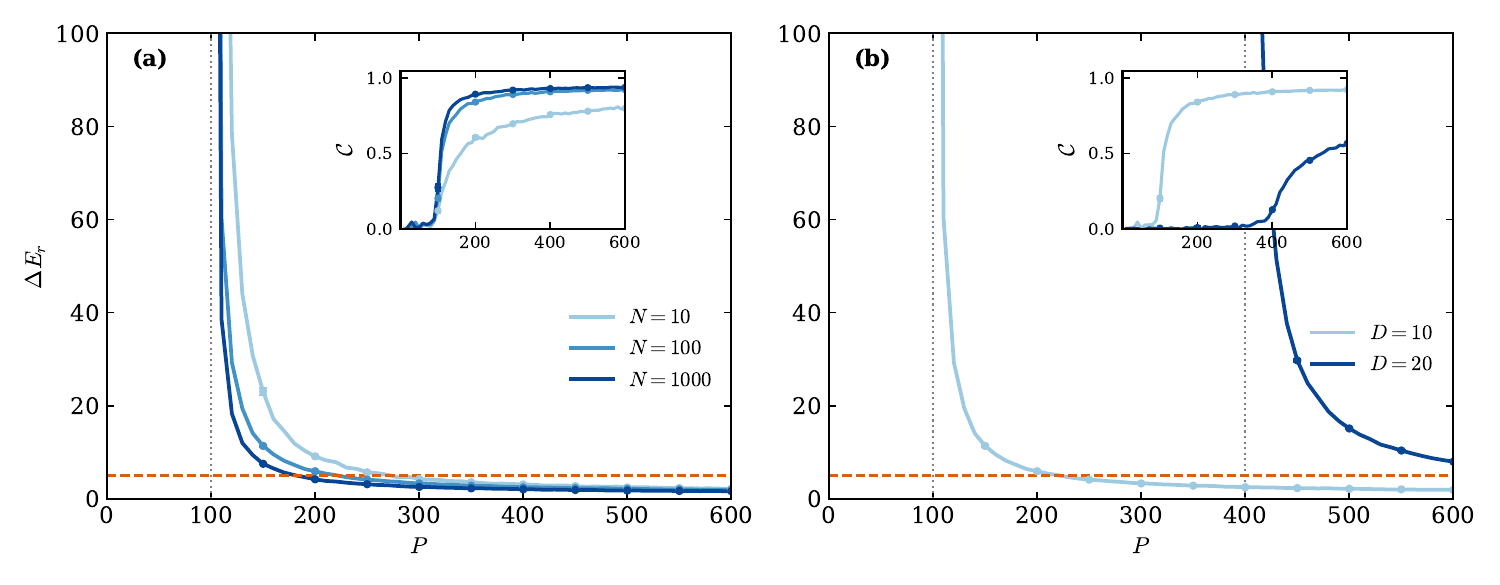}
    \caption{Dependence of the relative generalization gap $\Delta E_{r}$ and the contrast ratio $\mathcal{C}$ on task diversity $P$. We focus on $W_{21}=0$, and $\lambda=0$. The simulations of Langevin dynamics at $\beta=10^3$ and $\gamma=1$ are run up to $10^6$ steps (time step $\delta t=0.1$) for equilibrium. Note that the equilibrium can only be reached after the dotted lines, leading vanishing $\mathcal C$ and extremely large $\Delta E_r$ before the dotted lines. The main panels show the generalization gap, while the insets show the contrast ratio. Different color intensities correspond to different values of $N$ (or $D$). Lines are averages over $50$ training realizations. Error bars are shown only for specific values of $P$. Dotted lines mark $P=D^2$. (a) Results for different context lengths $N$ at fixed $D=10$. (b) Results for different input dimensions $D$ at fixed $N=100$. The dashed line of $\Delta E_r=5$ identifies a rough estimate of the value of $P$ separating memorization from generalization. }
        \label{fig5}
\end{figure}

Finally, we study the simplest setting introduced in Sec.~\ref{spe}. We consider the case of $\lambda=0$, which is shown in Fig.~\ref{fig5}. Note that in this case, $H$ is not invertable anymore when $P<D^2$. Therefore we apply the Langevin dynamics to plot Fig.~\ref{fig5}. For $\lambda=0$ and $P<D^2$, $H$ has $D^2-P$ exact zero eigenvalues. The data thus exert no force along these directions, so they diffuse freely and reach variance growing with time.
The transition occurs after the zero-eigenvalue fraction point, given this small value of $D$. The overall behavior is qualitatively the same with the case of $\lambda\neq0$. At $T=0$, the minimum-norm solution gives $E_t=0$ and therefore $\Delta E_r=\infty$.
At $P=D^2$, the smallest non-zero eigenvalue is about $10^{-7}$, which produces the peak observed in Fig.~\ref{fig5}.

\section{Conclusion}
In this work, we study the learning dynamics of single-head single layer linear in-context learning in a non-asymptotic setting, where the token dimension $D$, context length $N$ and the number of examples are all finite. A solvable learning dynamics equation is first obtained, where the eigenvalues of the Hessian matrix control how fast the weight parameters converge to the optimal state: the ridge solution of the problem. Next, we provide a data-dependent statistical mechanics description of the learning dynamics, from which the long-time limit of the training error and test error can be theoretically determined by the training data or test data. We find by tuning the value of $P$ (the number of training examples), the generalization gap decreases sharply, implying the presence of a memorization phase when the training linear-regression tasks are deficient; the memorization phase will be replaced by a generalization phase later as evidenced by a saturating behavior of the contrast ratio.

\begin{acknowledgments}
 This research was supported by the National Natural Science Foundation of China for Grant numbers 12475045, and Guangdong Provincial Key Laboratory of Magnetoelectric Physics and Devices (No. 2026B1212170049). We use GPT 6.1 sol for checking the results of this paper and providing advice to improve the manuscript quality. The authors take full responsibility for the scientific content of the paper.
\end{acknowledgments}

\section*{Code availability}
Codes to reproduce all results are deposited in our Github~\cite{codes}.


\end{document}